\def\year{2021}\relax
\documentclass[letterpaper]{article} % DO NOT CHANGE THIS
\usepackage{aaai21}  % DO NOT CHANGE THIS
\usepackage{soul}
\usepackage[utf8]{inputenc}
\usepackage{amsmath}
\usepackage{amsthm}
\usepackage{algorithm}
\usepackage{algorithmic}
\usepackage{amssymb}
\usepackage{multirow}
\usepackage{subfigure}
\usepackage{color}
\usepackage{booktabs}
\usepackage[switch]{lineno}
\usepackage{times}  % DO NOT CHANGE THIS
\usepackage{helvet} % DO NOT CHANGE THIS
\usepackage{courier}  % DO NOT CHANGE THIS
\usepackage[hyphens]{url}  % DO NOT CHANGE THIS
\usepackage{graphicx} % DO NOT CHANGE THIS
\usepackage{natbib}  % DO NOT CHANGE THIS AND DO NOT ADD ANY OPTIONS TO IT
\usepackage{caption} % DO NOT CHANGE THIS AND DO NOT ADD ANY OPTIONS TO IT
\title{Word Grounded Graph Convolutional Network}

\begin{document}

\author{
    Zhibin Lu\textsuperscript{\rm 1}\footnote{The first two authors contributed equally to this research.}
    Qianqian Xie\textsuperscript{\rm 2}\footnotemark[1]
    Benyou Wang\textsuperscript{\rm 3}
    Jian-yun Nie\textsuperscript{\rm 4}\\
}
\affiliations{
    \textsuperscript{\rm 1},\textsuperscript{\rm 4}University of Montreal, Montreal, Canada\\
    \textsuperscript{\rm 2}Department of Computer Science, University of Manchester, Manchester, United Kingdom\\
    \textsuperscript{\rm 3}University of Padova, Padova, Italy\\
    % \textsuperscript{\rm 4}University of Montreal, Montreal, Canada\\
    \textsuperscript{\rm 1}zhibin.lu@umontreal.ca
    \textsuperscript{\rm 2}qianqian.xie@manchester.ac.uk 
    \textsuperscript{\rm 3}wang@dei.unipd.it 
    \textsuperscript{\rm 4}nie@iro.umontreal.ca
}

\maketitle

\begin{abstract}
Graph Convolutional Networks (GCNs) have shown strong performance in learning text representations for various tasks such as text classification, due to its expressive power in modeling graph structure data (e.g., a literature citation network). Most existing GCNs are limited to deal with documents included in a pre-defined graph, i.e., it cannot be generalized to out-of-graph documents.
To address this issue, we propose to transform the document graph into a word graph, to decouple data samples (i.e., documents in training and test sets) and a GCN model by using a document-independent graph. Such word-level GCN  could therefore naturally inference out-of-graph documents in an inductive way.
The proposed Word-level Graph (WGraph) can not only implicitly learning word presentation with commonly-used word co-occurrences in corpora, but also incorporate extra global semantic dependency derived from inter-document relationships (e.g., literature citations). An inductive Word-grounded Graph Convolutional Network (WGCN) is proposed to learn word and document representations based on WGraph in a supervised manner.  Experiments \footnote{Our source code is available at~\url{ https://github.com/Louis-udm/Word-Grounded-Graph-Convolutional-Network.}} on text classification with and without citation networks evidence that the proposed WGCN model outperforms existing methods in terms of effectiveness and efficiency.
\end{abstract}
% for text representations learning. In the model, we propose to build the vocabulary graph via word co-occurrence and document relation information (i.e. literature citation ) to learn word representations. Based on it, the document representations can be achieved via document word feature transformation. Compare with methods learning on document level graph, WGCN model eliminates the dependence between the graph and documents, thus can inference representations for unseen documents based on learned word representations.
% Moreover, we theoretically analyze and prove the rationality and effectiveness of the graph building. We conducted comparative experiments on text and node classification tasks with multiple benchmark data sets. The experimental results show that our model is more compact and efficient compared with existing methods. Our method generalizes well to unseen documents along with improves the performance on node and text classification.
%\keywords{Graph Convolutional Networks, Vocabulary Graph, Text Classification}
\maketitle

%
%
%
%%%%%%%%%%%%%%%%%%%%%%%%%%%%%%%%%
%% Introduction %%
%%%%%%%%%%%%%%%%%%%%%%%%%%%%%%%%%
%
%
\section{Introduction}
%With the development of deep learning, 
Various deep neural networks, such as convolutional neural networks (CNNs) ~\cite{kim2014convolutional} and recurrent neural networks (RNNs) \cite{hochreiter1997long}, have been applied to %effectively and automatically 
learn textual representations. Both CNN and RNN based models can only capture the local semantic information among consecutive words, thus limiting their performance in modeling long documents, which contain long-range semantic dependency among non-consecutive words \cite{battaglia2018relational}.

Recently, a new type of network architecture named Graph Convolutional Network (GCN) \cite{battaglia2018relational,cai2018comprehensive,kipf2017semi,velivckovic2017gat} has been explored to learn text representations for specific tasks, such as text classification \cite{yao2019graph,peng2018large}. GCN models can effectively learn the node embedding of a given graph such as a literature citation network and a knowledge graph, by aggregating the global adjacent structural information into it. In these tasks, most existing GCN based methods  usually utilize the  given document correlation graphs \cite{kipf2017semi} or build  heterogeneous graphs containing word and document nodes \cite{yao2019graph}, to learn document representations. Compared with CNN and RNN based methods, they can capture global relations  between documents or model texts as order-free graph structure, thus are more powerful in learning document representations. However, the main issue of these methods is that they can only make inference about the documents  included in the pre-defined graph and can not be used for out-of-graph documents. This is  critical  when  test documents are integrated into the pre-defined graph during the training phase (e.g. in information stream applications). %especially in some cases, documents in the test set are not in the graph during training. 

% The training and test set must be included in the predefined graph before learning. Thus, they can't be generalized to out-of-graph documents.

To address this issue, we propose to transform the document-level graph into a word-level graph, to decouple data samples (i.e., documents in training and test sets) and GCN models with a document-independent graph. 
We build a  word-level graph (WGraph) to model semantic correlation between words, which is able to incorporate commonly-used word co-occurrence and extra semantic correlation derived from inter-document relationships.
Word co-occurrence, i.e. point-wise mutual information (PMI), is utilized to build the edges between two words in WGraph.
Extra document correlation information, if given, can be further effectively incorporated into WGraph via matrix transformation. 
%The edges between two words in WGraph are initially build by the word %co-occurrence, i.e, point-wise mutual information (PMI). If there is the extra %document correlation graph, they can be further updated by the inter-document %relationships via matrix transformation. 
Then, an inductive word grounded graph neural network (WGCN) is proposed to learn document representations based on the WGraph in a supervised manner. Different from previous learning methods on the given or constructed document-level graph, 
%our model can effectively combine both inter-document correlation information from the global word co-occurrence and the given graph in the word level-graph, 
our model leverages the word-level graph which combines the inter-document correlation information from the global word co-occurrence with the given graph, to learn the document representations and infer representations for unseen documents. It is worth noting that WGraph and WGCN are easily integrated by other broader inductive large language models \cite{ZhibinluGraphEmbedding}.

To summarize, our contributions are as follows:
\begin{itemize}
    \item We propose a novel inductive graph neural network called word grounded graph convolutional network (WGCN) for text representations learning.
    %\item We convert the transductive Text GCN to inductive learning mode, which is more elegant and efficient than SAGE~\cite{hamilton2017inductive} and FastGCN~\cite{chen2018fastgcn}.
    \item We propose to build a flexible word-level graph (WGraph) to model semantic correlation between words, which is able to incorporate word co-occurrence and extra semantic correlation.
    \item Experimental results of text and node classification on several benchmark data sets demonstrate the effectiveness and efficiency of our method.
\end{itemize}

%%%%%%%%%%%%%%%%%%%%%%%%%%%%%%%%%
%% Background and Related Work %%
%%%%%%%%%%%%%%%%%%%%%%%%%%%%%%%%%

\section{Related Work}
GCN has attracted much attention recently, it can be dated back to \cite{scarselli2008graph}, which was an extension of the recursive neural network (RNN) and random walk model, but limited to the traditional training manner of RNN.
\cite{li2015gated} revisited the model with modern optimization techniques and gated recurrent units. More recently, \cite{kipf2017semi} proposed a simplified graph convolutional neural network (GCN) via the first-order approximation of localized spectral graph convolutions, which achieved state-of-the-art performance on semi-supervised node classification.

Inspired by the effectiveness of GCN, many studies focused on extending it to text representation learning for specific tasks, such as semantic role labeling \cite{marcheggiani2017encoding}, text classification \cite{yao2019graph} and relation extraction \cite{zhang2018graph}. 
\cite{yao2019graph} proposed Text GCN model for text classification based on a document-word graph using word co-occurrence and document-word relations. 
\cite{linmei2019heterogeneous} further explored GCN on semi-supervised short text classification by integrating additional information (e.g., entities and topics). 
However, most of these methods cannot deal with  unseen documents: they require that all data samples (i.e., documents in train and test sets) are represented as the node in the predefined graph.
%However, due to coupling between data samples (i.e., documents in train and test sets) and graph, most of these methods can't inference documents not included in the predefined graph.

Several methods have been proposed to address this problem. \cite{huang2019text} proposed a text level graph neural network for text classification, which built a local graph for each text rather than a global corpus-level graph including all texts. 
Although the local graph relaxes the coupling between texts in graph building, it inevitably loses global semantic information.
%, it inevitably losses global semantic information in some extent.
\cite{velivckovic2017graph} proposed graph attention network (GAT), an attention-based architecture to perform node classification of graph-structured data, in which the attention mechanism makes the model applicable to inductive learning problems.
\cite{hamilton2017inductive} proposed GraphSAGE to learn an embedding function with good generalization ability for unseen nodes, via feature sampling of node neighborhoods.  
Rather than node neighborhoods, \cite{chen2018fastgcn} proposed FastGCN via important nodes sampling and revisiting the graph convolution as integral transforms of embedding functions. Different from their methods for learning an embedding function with generalization ability, our model tackles the problem by building a graph with low-level entities (e.g. words in vocabulary) with generalization ability, and utilizes them to infer representations for high-level entities (e.g. documents in corpus). 

We noticed that there are other  GCN methods using word-level graphs in various NLP tasks, such as word embedding learning \cite{vashishth2019incorporating}, text classification
\cite{peng2018large}, relation extraction
\cite{zhang2018graph}, machine translation
\cite{bastings2017graph} and question answering \cite{de2019question}. The difference lies in that our  word-level graph  is derived from a global document-level citation network, while the previous methods  usually build a local graph based on syntactic dependency graphs in a single document or sentence, i.e.  local graphs. We believe that GCN based on global citation relationships can capture complementary information with word co-occurrence. 
%\benyou{the word `ìnteresting'' seems subjective, in this case, I suggest to change ``interesting information shared across documents'' to ``complementary information with word co-occurrence.'''}

\section{Method}
\subsection{Graph Convolutional Network (GCN)}
In this section, we start from reviewing the graph convolutional network (GCN) \cite{kipf2017semi}, it is a powerful neural network for processing graph-structured data. A graph is defined as $G = (V, E)$, where $V (|V|=N)$ is a set of nodes and $E$ is a set of edges. GCN aims to learn node embeddings in the graph via information propagation in neighborhoods. To leverage the connectivity structure of the graph, assuming $A \in \mathbb{R}^{N \times N}$ is the adjacency matrix of $G$, each layer of GCN can be described as:
\begin{equation}
L^{(i+1)} = \sigma(\hat{A}L^{(i)}W_i)
\label{eq:1}
\end{equation}
where $\tilde{A}=D^{-\frac{1}{2}}(A+I)D^{-\frac{1}{2}}$ is the normalized symmetric adjacency matrix with self-loop for graph $G$, $D$ is the degree matrix of $A$, $I$ is an identity matrix, $L^{(i)}$ is the $i$th layer feature matrix for all nodes, $W_i$ is the weight matrix for $i$th layer, and $\sigma$ is an activation function. When applied to learn text representations, documents of the whole corpus are represented as nodes in a large graph, embeddings of them can be learned by stacking multi-layer GCN to capture multi-hop neighborhood information with such propagation process.

As with most other graph-based algorithms, the adjacency matrix is required to encode the pairwise relationships for both training and test data to induce their representations. However, in many applications, test data  should not be available during the training process.  Transductive learning approaches cannot cope with the problem. This requires the method to be inductive, i.e. to infer the representations of the unseen texts according to only the information of the  texts available in training. 

\subsection{Word Grounded Graph Convolutional Network (WGCN)}
\begin{figure*}[!hbt] 
  \vspace{0 cm}
  \centering
   \includegraphics[width=0.9\textwidth,height=3.2in]{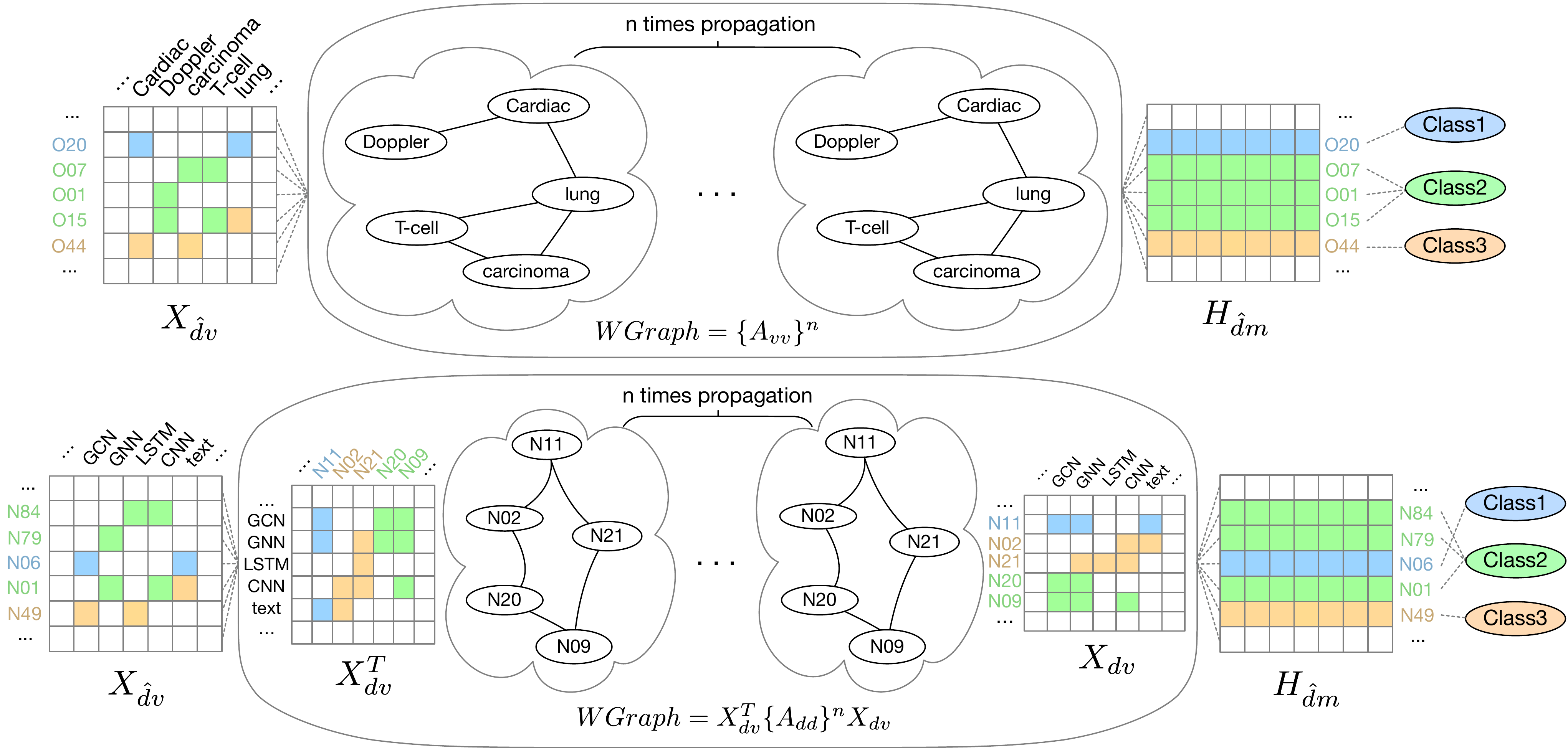}
  \caption[Schematic of WGCN model.]{Schematic of WGCN model and 2 word-level graph (WGraph) calculation methods. The first row shows the word convolutional via $n$ times propagation on the WGraph $A_{vv}$ without the document correlation graph, the second row shows the word convolutional via $n$ times propagation on the document correlation graph $A_{dd}$. The subscript $d$ indicates the training documents, while $\hat{d}$ indicates another document set, which can be either training or test documents.}
  \label{fig:model}
\end{figure*}

To infer out-of-graph samples, we propose the word grounded GCN to make it feasible for inductive inference. 
%The idea is depicted in Figure \ref{fig:gcn-vs-WGCN} and Figure \ref{fig:model},
The idea is depicted in Figure \ref{fig:model}, in which a document graph is translated into a word-level graph. In this subsection, we start from the following two key problems: 1) How to build a document-independent graph without losing the original document correlation information?
2) How to effectively incorporate the word co-occurrence and document adjacency graph to learn document representations?

% \begin{figure*}[!htp]
%   \centering
%   \includegraphics[width=100 mm
%   %\textwidth 
%   %height=120 mm
%   ]{figure/gcn-vs-wgcn.pdf}
%   \caption[Illustration of GCN and WGCN for one document.]{Illustration of GCN and WGCN for one document. In GCN, the representation $H2$ of document 2 is aggregated by all other documents (doc1, doc3, doc4) in the document-level graph. In WGCN, the representation $H2$ of document 2 is aggregated by all words of document 2 (word2, word3, word4) and other words connected (word1) in the word-level graph (WGraph).
% }
%   \label{fig:gcn-vs-WGCN}
% \end{figure*}

\textbf{Building Word-level Graph}: Given a corpus $D (d=|D|)$ with vocabulary size $v=|V|$, we build a flexible word-level graph (WGraph), whose nodes are all words in $V$.
We consider two cases: \textit{ documents are mutually independent} and  \textit{documents are correlated}.

\textit{Documents are  mutually independent.} Typically,  We can directly represent a word by observing whether it appears (or possibly with the term frequency) in all documents, i.e., $\vec{w_i}^T = [x_{i1},x_{i2},\cdots, x_{id}]$. Let us denote $X_{dv}$ as  a document-word matrix  depending on a corpus $D$ in which $i$-th column of $X_{dv}$ (denoted as $\vec{w_i}$) could be the presentation of  $i$-th word.
% Technically, a word can be represented with a feature vector with the same size as the total document set.
%\begin{equation}
%    \vec{w_i} = [x_{i1},x_{i2},\cdots, x_{id}]
%    \label{eq:7}
%\end{equation}
A correlation between two words can be calculated by a dot product between two word feature vectors i.e., $c_{ij}=\vec{w_i}^T\vec{w_j} $. 
Namely, a correlations between two arbitrary words can be defined as:
\begin{align}
A_{vv} = X_{dv}^{\mathrm{T}}X_{dv} \in \mathbb{R}^{v \times v}
\label{eq-pmi-is-xx}
\end{align}
resulting in a matrix with dimensions $v \times v$. This is also a way to express the co-occurrence of words in the documents. With proper normalization, the above between-word  correlation matrix $A_{vv}$ can be transformed into an adjacency matrix to build a graph. For example, a Point-wise Mutual Information (PMI, or normalized point-wise mutual information (NPMI)) matrix is one of the typical  word correlation matrices to use \cite{yao2019graph}.

\textit{Documents are correlated}. By removing the prior assumption of  independence, i.e. documents are correlated with some relations, for example,  literature citation relations. We can define a more general  word correlation matrix. Technically, the $i$-th word feature with $n$-order inter-document relationships is defined as follows:
\begin{equation}
\begin{aligned}
    \vec{w_i}^{(n)} &= \left([x_{i1},x_{i2},\cdots, x_{id}] \underbrace{A_{dd}\cdots A_{dd}}_\text{n times}\right)^T \\
    &= \left(\vec{w_i}^T {\{A_{dd}\}}^n\right)^T
\end{aligned}
\label{eq:2}
\end{equation}
$n$ determines how many hops of citation relationships the word representation  adopts. A bigger $n$ of  $\vec{w_i}^{(n)}$ indicates a longer hop citation will be leveraged.
Based on the order of inter-document relationship used for  word representations, a  general definition of word correlations is as follows:
\begin{equation}
    A^{(m,n)}_{vv} = X_{dv}^T {\{A_{dd}\}}^m {\{{A_{dd}}^T\}^n}X_{dv} 
    \label{eq:3}
\end{equation}
where $A^{(m,n)}_{vv}$ denotes the correlations between word representations utilizing $m$-order and $n$-order inter-document relationships.
%Note that $A_{dd}$ will not necessarily be equal to $A_{dd}^T$ if it is not symmetric.
% More specifically, Eq \ref{} is a special case of Equation \ref{eq:9}, in which cases $m=n=0$ and assuming documents are independent. 
%Depending on the order of inter-document relationship a real-world application needs, one can empirically choose $m$ and $n$ to fit the application. 

According to previous methods of GCN \cite{kipf2017semi}, $A_{dd}$ will be transformed into a symmetric Laplacian matrix \footnote{In rest of the paper, we denote the transformed matrix as  $A_{dd}$ since there is no risk of confusion.}, namely, $A_{dd}= A_{dd} ^T$. Thus we can rewrite the formula in Equation \ref{eq:3} as follows:
\begin{equation}
    A^{(k)}_{vv} = X_{dv}^T { \{ A_{dd} \} }^k X_{dv} 
    \label{eq:4}
\end{equation}
where $k = m+n$. In the document-independent WGraph, we effectively unify both word co-occurrence ($X^T_{dv}X_{dv}$) and the inter-document relation information ($A_{dd}$), to make them complement each other to improve the document representation learning. In this convolution operation, the first word representation $X^T_{dv}$ will aggregate the relationship of the citation relationship $A_{dd}$, and then multiplied by the second $X_{dv}$ to obtain the final correlation, which means that the word co-occurrence matrix is aggregated by the citation relationship. %Considering the case of giving a citation network, we believe that the WGraph can capture the relation between two documents sharing most words with the word co-occurrence, which are not included in the given citation network. 

\textbf{Word Graph Convolution}: We propose a word grounded graph convolutional network (WGCN) to embed WGraph for learning document representations. 

First, we apply  Equation \ref{eq:1} on the adjacency matrix $A_{vv}$ to learn the representations of words:

\begin{equation}
    H_{vm} = \sigma(\{A_{vv}\}^nX_{v}W_0)
    \label{eq:Hvm}
\end{equation}
where $H_{vm} \in \mathbb{R}^{\hat{v} \times m}$ is transformed word representations with each row corresponding to a word, $\sigma$ is an activation function, and $W_0 \in \mathbb{R}^{v \times m}$ is a weight matrix. $A_{vv}$ is a word-wise correlation matrix like in Equation \ref{eq-pmi-is-xx}.
Since we have already involved the raw word feature matrix   $X^T_{dv}$  to calculate $A_{vv}$ in Equation \ref{eq-pmi-is-xx}, here we simply set word feature matrix $X_{v} = I \in \mathbb{R}^{v \times v}$ as an identity matrix which means every word is represented as a one-hot vector to represent itself. 

To predict the label of documents during testing phase, we first get the documents'
raw feature vector which could be a distribution on the whole vocabulary, e.g., term frequency vector for a document, denoted as $X_{\hat{d}v}$. 
Then we get the document representation by   multiplying a document-word feature representation $X_{\hat{d}v}$ and the word representation vector $H_{vm}$ as below:
\begin{equation}
\begin{aligned}
  H_{\hat{d}m} &= X_{\hat{d}v}H_{vm}  \\
  &= X_{\hat{d}v} \sigma(\{A_{vv}\}^nX_{v}W_0)
\label{eq:5}  
\end{aligned}
\end{equation}
where $H_{\hat{d}m}\in \mathbb{R}^{\hat{d} \times m}$  and each row in $H_{\hat{d}m}$ represents a specific document.  $X_{\hat{d}v} \in \mathbb{R}^{\hat{d} \times v}$ is the feature matrix of input documents. Note that $X_{\hat{d}v}$ is different from $X_{dv}$ since $X_{dv}$ is fixed as the document-word feature matrix of training documents, while $X_{\hat{d}v}$ can be the feature matrix of training or test data, which enables our model to infer representations for  unseen documents.

Similar to \cite{simplifyGCN}, we apply a nonlinear activation function after $n$ times message propagation to simplify the model structure, resulting in a linear structure with the $n$-th power adjacency matrix as $\{A_{vv}\}^n$ and a single weight matrix $W$.
After learning the word node embeddings in the WGraph, the document-word feature matrix of the input data $X_{\hat{d}v}$ is utilized to generate final document representations $H_{\hat{d}m}$.

As for a scenario with the provided document correlation graph, the propagation rule for the adjacency matrix $A_{vv}$ is different:
\begin{equation}
  H_{\hat{d}m} = X_{\hat{d}v} \sigma((X_{dv}^{\mathrm{T}}\{A_{dd}\}^n X_{dv})X_vW_0)
\label{eq:6}  
\end{equation}
where $A_{dd} \in \mathbb{R}^{d \times d} $ is the given document correlation graph, and $X_{v} = I \in \mathbb{R}^{v \times v}$ is also an identity matrix.
Only the $n$-th power of $A_{dd}$ is utilized to perform the message propagation while $X_{dv}$ does not participate in the propagation. This is done to fully leverage the pairwise information between documents in the predefined graph, for example, multiple-hop citations. Overall, our model is applicable for both without and with the document-level graph scenarios, in a sense a document-independent assumption will be a special case when $A_{vv}$ in Eq. \ref{eq:6} becomes an identify matrix (i.e., a matrix with diagonal elements being ones and zeros elsewhere).

For the classification task, we construct a word graph using the words in the documents. The words are connected according to their NPMI determined from the processing of the document collection. then we learn a document representation  by WGCN, which is then  fed into a simple MLP layer, followed by the softmax classifier to predict target labels. The loss function can be defined as:
\begin{equation}
\begin{split}
  \hat{Y}_{\hat{d}c} &= \textrm{Softmax}(\textrm{ReLu}(H_{\hat{d}m}) W_1 + b_1)\\
  \mathcal{L} &= -\sum_{i \in \hat{d}}\sum_{j \in c} Y_{i,j}\ln \hat{Y}_{i,j}
\label{eq:7}  
\end{split}
\end{equation}
where $W_1 \in \mathbb{R}^{m \times c}$ is the weight matrix, $b_1 \in \mathbb{R}^{c}$ is the bias, $Y$ is the label indicator matrix, $c$ is the number of classes.

\subsection{Comparison with Existing Methods}

In this subsection, we provide an analysis on the connection and difference of our method with existing classical methods. 

First, we start to revisit our model without a given document-level graph to the following formulation:
\begin{equation}
\begin{split}
  H_{\hat{d}m} &=X_{\hat{d}v} \{A_{vv}\}^n X_{v}W_0\\
  &= X_{\hat{d}v} X_{dv}^{\mathrm{T}} \{X_{dv}X^T_{dv}\}^{n-1} X_{dv} X_{v}W_0\\
  &= A_{\hat{d}d}\{\hat{A}_{dd}\}^{n-1} X_{dv}W_0
\label{eq:8}
\end{split}
\end{equation}
where $A_{\hat{d}d}$ is the similarity matrix between the input documents $\hat{d}$ and all training documents $d$ in corpus, $\hat{A}_{dd}=X_{dv}X_{dv}^{\mathrm{T}}$. In the formulation, we collapse the identity matrix $X_v$ and make a linearization of our model via removing the the only non-linear activation function similar to \cite{simplifyGCN}, which had proven the useless of non-linearization in information propagation. 

For comparison, the linearized Text GCN model \cite{yao2019graph} according to \cite{simplifyGCN} can be derived as:
\begin{equation}
\begin{split}
  H_{dm} &=\{\hat{A}\}^nX_{dv}W_0\\
  &= {\begin{pmatrix}
A_{dd} & X_{dv} \\
X_{dv}^{\mathrm{T}} & A_{vv}
\end{pmatrix} }^{n}X_{dv}W_0\\
& = {\begin{pmatrix}
I & X_{dv} \\
X_{dv}^{\mathrm{T}} & A_{vv}
\end{pmatrix} }^{n}X_{dv}W_0
\label{eq:9}
\end{split}
\end{equation}
where $\hat{A}$ is a heterogeneous graph with document and word nodes $d+v$.
Moreover, the linearized GCN \cite{kipf2017semi} model according to \cite{simplifyGCN} is:
\begin{equation}
\begin{split}
  H_{dm} &= \{A_{dd}\}^n X_{dv}W\\
\label{eq:10}  
\end{split}
\end{equation}
From Equation \ref{eq:8},\ref{eq:9} and \ref{eq:10}, we can found that our linearized model basically has the similar convolutional formula with Text GCN and GCN models except for the adjacency matrix for information propagation. In Text GCN model, both word co-occurrence $A_{vv}$ and document-word relations $X_{dv}$ are included in the graph, while our model only keeps the former. In our method, $X_{dv}$ is decoupled from graph and directly utilized to model the document-word relations. Moreover, from Equation \ref{eq:8}, we can explain our model from another perspective, that our model can learn representations of available documents on the inter-document graph $\hat{A}_{dd}$ derived from $X_{dv}$ with the same convolutional layer as GCN model, then effectively inference representations for unseen documents via the similarity information between available and unseen documents.

In similar way, our model with a given document-level graph can also be linearized to the following formulation:
\begin{equation}
\begin{split}
  H_{\hat{d}m} &= X_{\hat{d}v} (X_{dv}^{\mathrm{T}}\{A_{dd}\}^n X_{dv})X_vW\\
  &= A_{\hat{d}d}\{A_{dd}\}^n X_{dv}W
\label{eq:11}  
\end{split}
\end{equation}
From Equation \ref{eq:10} and \ref{eq:11}, it is clear that our model has a same convolutional operation as GCN model on the document-level graph, along with the effective inductive inference mechanism via the inter-document relations $A_{\hat{d}d}$ between unseen and original documents.

\section{Experiments}
We conduct experiments on text classification with and without citation networks to evaluate our model. There are two major questions we aim to study:
\begin{enumerate}
    \item  How effective is our method in text classification without citation networks compared with existing approaches?
    \item How effective is our method in text classification with citation networks the compared with existing approaches?
\end{enumerate}

\subsection{Text Classification without Citation Networks}
% We first describe our experimental settings and report the overall results in text classification to answer the first question when pairwise document information is not given.
We first assume that the documents are mutually independent, i.e. in the ordinary text classification datasets, there is no explicit relationship between the documents that can be used. We use the same datasets as in \cite{yao2019graph}, including 20-Newsgroups (20NG), Ohsumed, R52 and R8 of Reuters, and Movie Review (MR).
The overview of the five datasets is depicted in Table \ref{tab:statistics}.
\begin{table}[t]
    \scriptsize
    \centering
    \renewcommand{\arraystretch}{1.2}
    \caption{Summary statistics of five datasets~\protect\cite{yao2019graph}}
    \setlength{\tabcolsep}{1.5mm}{
    \begin{tabular}{c|ccccccc}
    \hline
    \bf{Dataset}& \bf{Docs}	& \bf{Train}& \bf{Test}& \bf{Words} & \bf{Nodes}& \bf{Classes} & \bf{Average Len} \\
    \hline
    20NG & 18,846 & 11,314 & 7,532 & 42,757 & 42,757 & 20  & 221.26\\
     R8 & 7,674 &5,485 & 2,189 & 7,688 & 7,688 & 8 & 65.72\\
     R52& 9,100 & 6,532	 & 2,568 & 8,892& 8,892 & 52 & 69.82\\
    Ohsumed& 7,400 & 3,357 & 4,043 & 14,157 & 14,157 & 23 & 135.82\\
    MR& 10,662 & 7,108 & 3,554 & 18,764 & 18,764 & 2 &20.39\\
    \hline
    \end{tabular}}
    \label{tab:statistics}
    \end{table}
    
%\textbf{Baselines}: We compare our methods with the following methods: 1) \textbf{CNN} \cite{kim2014convolutional}: a CNN-based method; 2) \textbf{LSTM} \cite{liu2016recurrent}: a LSTM-based method; 3) \textbf{Graph-CNN} \cite{defferrard2016convolutional}: a CNN-based method with word embedding similarity graph; 4) \textbf{Text-GCN} \cite{yao2019graph}: a GCN-based method with a global graph for all documents and words; 5) \textbf{Text-GNN} \cite{huang2019text}: a GNN-based method with a graph for each document; 6) \textbf{Fast-GCN} \cite{chen2018fastgcn}: a GCN-based method enhanced with importance sampling. 

\textbf{Baselines}: We compare our methods with the following methods: 
\begin{itemize}
\item \textbf{CNN} \cite{kim2014convolutional}: a CNN-based method; 
\item \textbf{LSTM} \cite{liu2016recurrent}: a LSTM-based method;
\item \textbf{Graph-CNN} \cite{defferrard2016convolutional}: a CNN-based method with word embedding similarity graph; 
\item \textbf{Text-GCN} \cite{yao2019graph}: a GCN-based method with a global graph for all documents and words; 
\item \textbf{Text-GNN} \cite{huang2019text}: a GNN-based method with a graph for each document. In this case, the author did not publish the source code, so we directly quote the performance in the paper; 
\item \textbf{Fast-GCN} \cite{chen2018fastgcn}: a GCN-based method enhanced with importance sampling. 
\end{itemize}

\textbf{Experimental Settings}: We set the hidden size of the convolutional layer among $\{100, 200, 250, 500, 550\}$, the dropout rate to 0.6, the learning rate among $\{0.016, 0.018,0.02\}$, the window size among $\{10, 20, 25, 50\}$, the weight decay among $\{0,5e-5\}$, the maximum training epochs with Adam to 800, the early stop epoch to 10. The parameter settings in all baseline models are the same as in \cite{yao2019graph,chen2018fastgcn,huang2019text}.

\textbf{Performance}: As shown in Table \ref{tab:results}, our proposed method achieves the best performance in all datasets, which demonstrates the effectiveness of our method. When comparing with Text-GCN, our method yields around 1\% improvement in the accuracy. Recall that Text-GCN builds a large graph including documents (and test documents) and word nodes, while our model builds a graph containing only word nodes. The results show that our graph can generate comparable or slightly better document representations to those generated via Text-GCN, based on the word graph where no test documents are included.
%reproduce document representations from word graph which are of at least comparable to (slightly better than) that deduced in Text-GCN graph.
%However, we do not have to include the test documents into the graph as Text-GCN. 
%In fact, in addition to the occurrence information of terms in documents, another important piece of information is the word-occurrence information captured in our graph, which helps to construct document representations without the limitation of using only the words in the document. 
%to improve the overall performance while decoupling documents from the graph. 

Both Fast-GCN and Text-GCN do not consider global correlations between words, which can be captured in our graph. %When enhanced with importance sampling over samples or building a local graph for each document, Fast-GCN and Text-GCN inevitably lack the ability 
They are unable to utilize global word correlation information. The comparison between our method and these methods show the importance of capturing  global word correlations in a graph.
%In contrast to them, we propose to decouple the documents from the graph, which allows us to keep the global correlations of words in the graph to achieve better performance.

Note that the results of the graph-based methods tend to outperform other non-graph based methods, such as CNN, LSTM, and fastText. The reason is that the latter cannot leverage the information from a graph structure, while the graph-based methods can learn more expressive representations for nodes considering their neighborhoods. 

In Table \ref{tab:A-results}, we further use different order of $A_{vv}$ on the datasets of R52 and R8 to test the accuracy of our method. We choose these two datasets because other datasets will cause out of GPU memory when calculating $A^2_{vv}$ and $A^3_{vv}$. On both datasets, our method achieves the best performance when the 1-order neighborhood information in $A_{vv}$ is incorporated. With $0$-order $A_{vv}$, our method degrades to a two-layer MLP model with no neighborhood information incorporated. On the other hand, using higher-order information may over-propagate information to moderately related documents and noise can be introduced.

\begin{table*}[t]
\small
    \centering
    \renewcommand{\arraystretch}{1.2}
    \caption{Test Accuracy on document classification task for 10 times running and report mean $\pm$ standard deviation. WGCN with one convolutional layer steadily outperforms other methods on 20NG, MR, R52 and R8 based on student t-test ($p-value <0.05$). \textbf{OOM:} Out of GPU Memory.}
    \begin{tabular}{l|c|c|c|c|c}
    \hline
    \bf{Model}      & \bf{20NG}	          & \bf{MR}             & \bf{Ohsumed}            & \bf{R52}        & \bf{R8}\\ 
    \hline
    CNN & 0.8215 $\pm$ 0.0052 & 0.7775 $\pm$ 0.0007 & 0.5844 $\pm$ 0.0106 & 0.8759 $\pm$ 0.0048 & 0.9571 $\pm$ 0.0052 \\
    LSTM & 0.7318 $\pm$ 0.0185 & 0.7768 $\pm$ 0.0086 & 0.4927 $\pm$ 0.0107 & 0.9054 $\pm$ 0.0091 & 0.9631 $\pm$ 0.0033 \\
    Graph-CNN & 0.8142 $\pm$ 0.0032 & 0.7722 $\pm$ 0.0027 & 0.6386 $\pm$ 0.0053 & 0.9275 $\pm$ 0.0023 & 0.9699 $\pm$ 0.0012 \\
    Fast-GCN     & \textbf{OOM} & 0.7510 $\pm$ 0.0021 & 0.5441 $\pm$ 0.0081 & 0.8515 $\pm$ 0.0045 & 0.9538 $\pm$ 0.0036 \\
   	Text-GCN     & 0.8634 $\pm$ 0.0009 & 0.7674 $\pm$ 0.0020 & 0.6836 $\pm$ 0.0056 & 0.9356 $\pm$ 0.0018 & 0.9707 $\pm$ 0.0010 \\
   	Text-GNN     & -  & - & 0.6940 $\pm$ 0.0060 & 0.9460 $\pm$ 0.0030 & 0.9780 $\pm$ 0.0020 \\
	WGCN  & \textbf{0.8885} $\pm$ 0.0012 &  \textbf{0.7794} $\pm$ 0.0010  & \textbf{0.6962} $\pm$ 0.0024 & \textbf{0.9486} $\pm$ 0.0009 & \textbf{0.9790} $\pm$ 0.0014 \\
    \hline
    \end{tabular}
    \label{tab:results}
\end{table*}

\begin{table}[!hbt]\footnotesize
    \centering
    \renewcommand{\arraystretch}{1.2}
    \caption{Test Accuracy on document classification task for 10 times running using different order of $A_{vv}$, i.e. $A_{vv}^k, k\in(0,1,2,3)$. }
    \begin{tabular}{c|c|c}
    \hline
    \bf{Model}      & \bf{R52}	          & \bf{R8}  \\ 
    \hline
	$A_{vv}^0$ & 0.9172 $\pm$ 0.0018 &0.9509 $\pm$ 0.0013 \\ 
	$A_{vv}^1$ &  \textbf{0.9486} $\pm$ 0.0009 & \textbf{0.9785} $\pm$ 0.0014 \\
	$A_{vv}^2$ & 0.9062 $\pm$ 0.0012 &  0.9671 $\pm$ 0.0015   \\
	$A_{vv}^3$ & 0.7558 $\pm$ 0.0012 &  0.8995 $\pm$ 0.0018   \\
    \hline
    \end{tabular}
    \label{tab:A-results}
\end{table}

\begin{figure}[!hbt] 
  \vspace{0 cm}
  \centering
  
  \subfigure[hidden dimension]{
%   \begin{minipage}[c]{0.23\textwidth}
  \includegraphics[width=0.35\textwidth,height=1.6in]{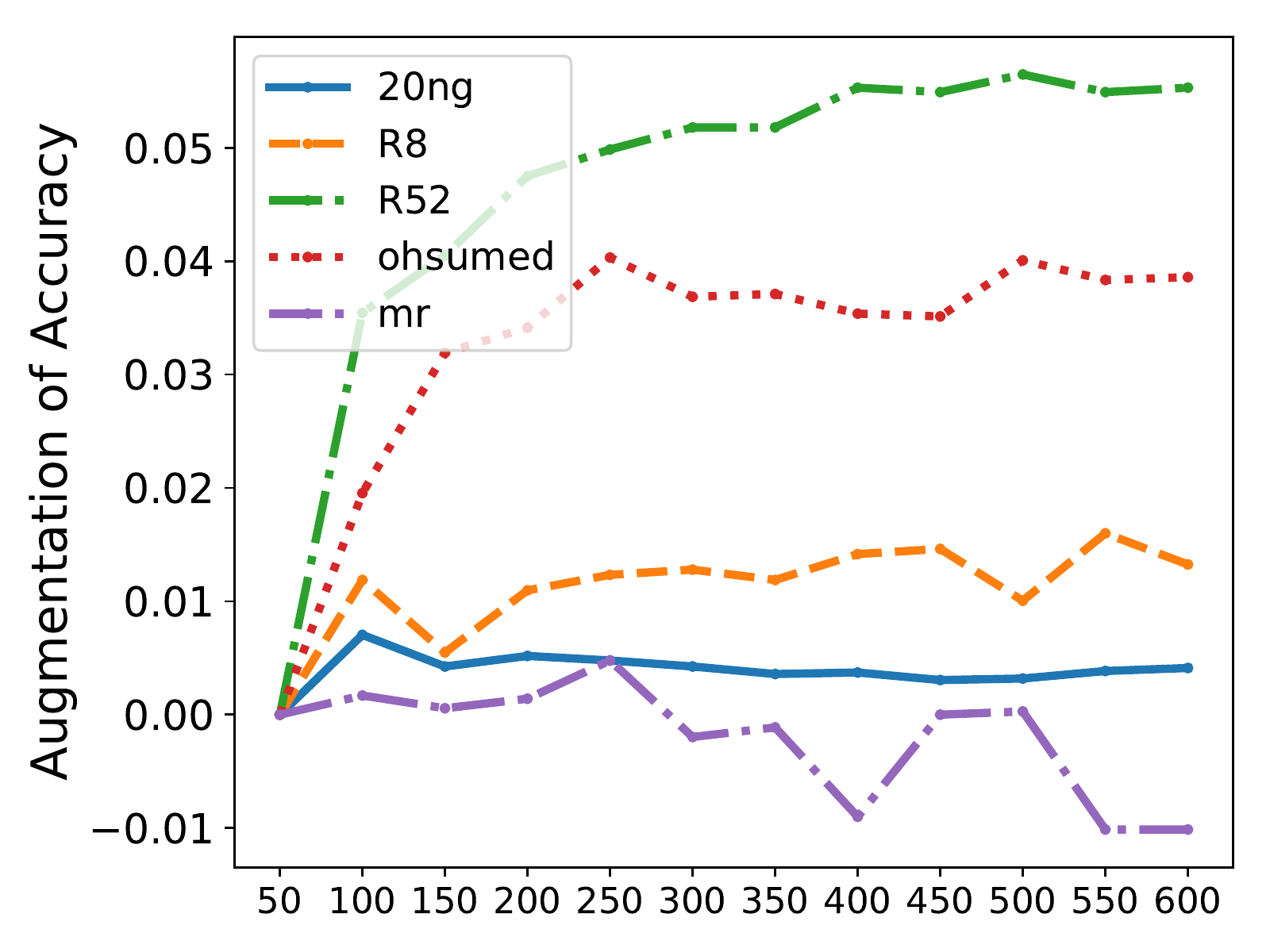}
%   \end{minipage}
}

  \subfigure[window size]{
%   \begin{minipage}[c]{0.23\textwidth}
   \includegraphics[width=0.35\textwidth,height=1.6in]{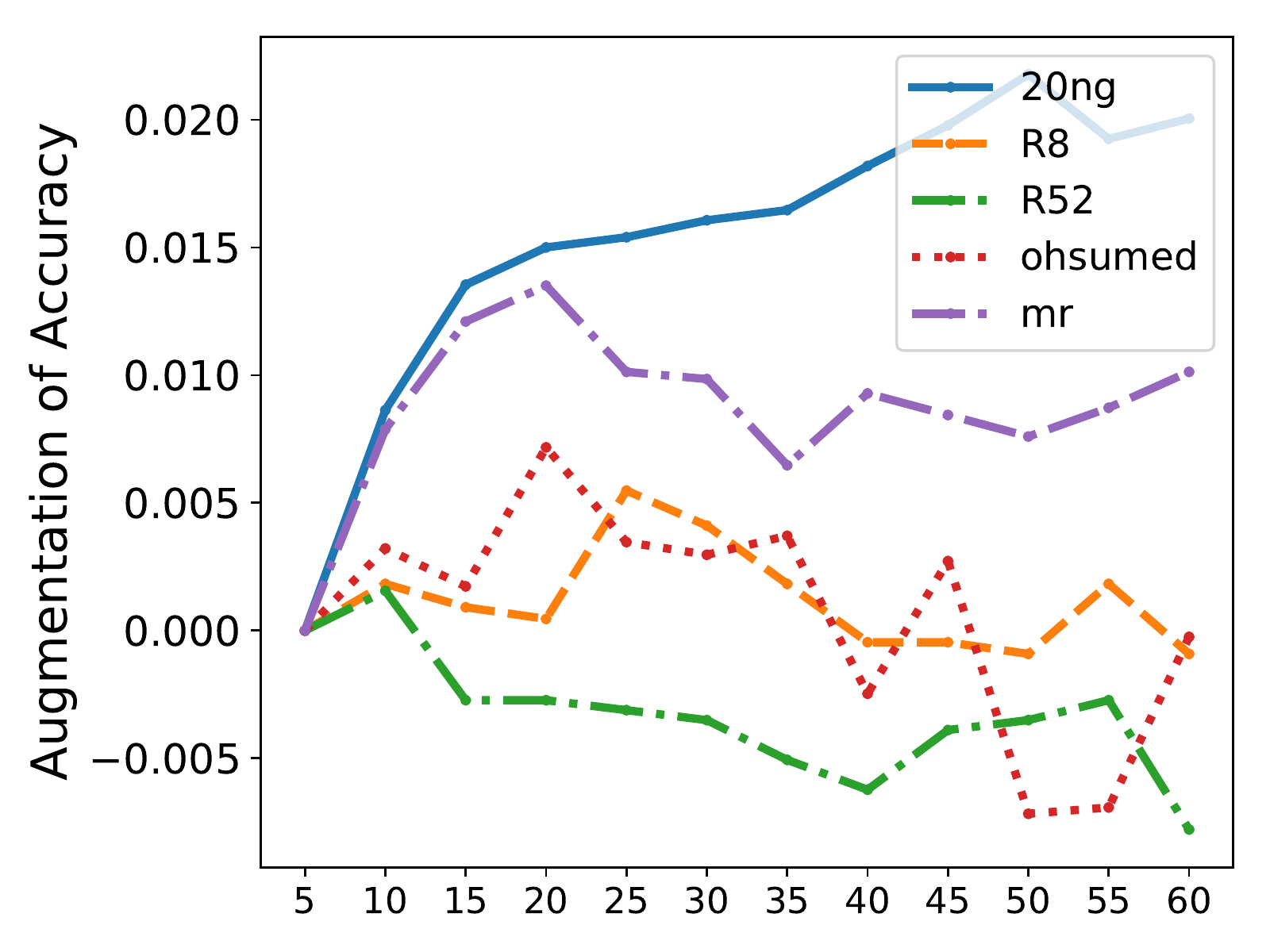}
%   \end{minipage}
}

  \caption{The augmentation of test accuracy with WGCN under different hidden dimension and sliding window size.}
  \label{fig:parameter}
\end{figure}

\textbf{Parameter Sensitivity:} Figure \ref{fig:parameter} shows the accuracy with our model on five datasets using different hidden dimension of the convolutional layer and sliding window size in PMI calculation. We can see that:
1) On R8, R52, and Ohsumed, 
%1) when the hidden dimension is smaller than $250$, on datasets R8, R52, Ohsumed and MR,
the test accuracy improves when the hidden dimension increases up to $250$, then it drops slowly.
%When the hidden dimension is larger than $250$, in contrast, the test accuracy increases slowly on datasets R8, R52, Ohsumed, but decreases on MR dataset.
2) For R8, MR and 20NG datasets, varying hidden dimension has limited influence on the test accuracy.
Overall, it shows that the hidden dimension can neither be too small to embed enough information for propagation in the graph, nor too large to focus on important information for classification.
%Overall, these results show that too small hidden dimension may not have enough representations ability for information propagation in the graph, while a too large hidden dimension may capture too much details that are not important for the classification task, thus yielding lower performance.

Similarly, on R8, R52, Ohsumed and MR datasets, our methods yield the best result when the window size is around 20, while smaller or larger window size can capture insufficient or noisy information. 
%Similar to the hidden dimension, on R8, R52, Ohsumed and MR datasets, the test accuracy has the highest value when the window size is around 20. Too small or too large window size may generate insufficient word co-occurrence or introduce noise information, thus yielding lower performance.
For 20NG dataset, the test accuracy consistently increases with the growing of window size.
A larger window size may help capturing sufficient word correlation information with long documents, yielding better performance.
%We observe that this may be due to the larger average document length than other datasets. A larger window size may help generating sufficient word correlation information in this collection, thus yielding better performance.

The above experiments show that the hidden dimension and window size should be chosen according to each test collection. It may not be good to use the same setting for different datasets. 

\textbf{Efficiency}: We further demonstrate that our method is significantly faster and has lower memory consumption compared with other GCN-based methods, with comparable and even better classification performance. The time complexity and space complexity of a one-layer GCN depend on $A_{dd}$, i.e. $O(d^2)$, while its depend on $A_{vv}$ for a one-order WGCN, i.e. $O(v^2)$. The difference between $O(d^2)$ and $O(v^2)$ is that in most cases, the vocabulary is relatively fixed, therefore $|V|$ do not grow or grows very slowly, while the dataset size ($|D|$) may be very large, and the disadvantage of GCN is that all documents must be loaded regardless of training phase or evaluation phase. We show in Table \ref{tab:timeComplexity} the time cost and the GPU memory  cost of different models on one epoch of training. We can see that WGCN has obvious advantages on both costs. In terms of time cost, Fast GCN is on average 14.6 times longer than WGCN, and Text-GCN is on average 5.1 times longer than WGCN. In terms of space cost, Fast GCN is on average 14.0 times larger than WGCN, and Text GCN is 4.5 times larger than WGCN. Notice that on the large dataset 20NG,  Fast-GCN and Text-GCN models cannot be trained with GPU due to out of memory problem. Different from Fast-GCN and Text-GCN, our model removes useless nonlinear function and collapses multi-weight matrices into a single weight matrix among consecutive GCN layers. This  significantly reduces both consumptions in time and memory. 

\begin{table}[t]\scriptsize
    \centering
    \renewcommand{\arraystretch}{1.2}
    \caption{Time cost (ms) and GPU memory space cost (MiB) on one epoch of training under CUDA. (GPU: Nvidia Tesla K40c, CPU: Intel(R) 8 Core(TM) i7-2600K CPU @ 3.40GHz)}
     \setlength{\tabcolsep}{1.4mm}{
    \begin{tabular}{l|c|c|c|c|c}
    \hline
    \bf{Model}      & \bf{20NG}	          & \bf{MR}             & \bf{Ohsumed}            & \bf{R52}        & \bf{R8}\\ 
    \hline
    & Time / Space & Time / Space & Time / Space & Time / Space & Time / Space \\
   	Fast-GCN     &\textbf{N/A} / \textbf{OOM}               &9,036 / 8,841     &4,016 / 8,841   &4,826 / 8,585   &3,602 / 4,489     \\
    Text-GCN     &\textbf{N/A} / \textbf{OOM} &1,060 / 3,995     &4,303 / 2,960   &2,365 / 2,119	&1,869 / 1,613 \\
    WGCN     &6,435 / 2,556  &458 / 567	    &1,905 / 959    &1,020 / 717	&812 / 638 \\
    %WGCN(NPMI)    &2,699 / 1,330  &286 / 522      &739 / 696    &425 / 501    &351 / 479 \\
    \hline
    \end{tabular}}
    \label{tab:timeComplexity}
\end{table}

\subsection{Text classification with Citation Networks}
In this subsection, we present and analyze the results on the datasets in which the documents are correlated (i.e. a predefined document-level graph -- the citation networks).

\textbf{Datasets and Baselines}: We use the same datasets as in \cite{kipf2017semi}, including Citeseer, and Pubmed citation datasets, in which a graph with the inter-document citation information is available.
The overview of datasets is depicted in Table \ref{tab:citation-statistics}.
We compare our method with the following methods: 
\begin{enumerate}
    \item  \textbf{GCN} \cite{kipf2017semi}: the original transductive GCN model,
    \item \textbf{Fast-GCN} \cite{chen2018fastgcn}: an inductive GCN method which adopts importance sampling to do inference for unseen samples,
    \item \textbf{GIN} \cite{howpowerfulGCN}: the Graph Isomorphism Network (GIN) which has a large discriminative power compared with GCNs, 
    \item \textbf{SGC} \cite{simplifyGCN}: a simplified linear GCN model.
\end{enumerate} 
\begin{table}[t]
    \small
    \centering
    \renewcommand{\arraystretch}{1.2}
    \caption{Statistics of the citation network datasets~\protect\cite{simplifyGCN}}
    \begin{tabular}{c|cccc}
    \hline
    \bf{Dataset}& \bf{Nodes}& \bf{Edges}& \bf{Classes}& \bf{Train/Dev/Test Nodes} \\
    \hline
    Citeseer & 3,327 & 4,732 & 6 & 120/500/1,000\\
    Pubmed& 19,717 & 44,338 & 3 & 60/500/1,000\\
    \hline
    \end{tabular}
    \label{tab:citation-statistics}
    \end{table}
    
\textbf{Experimental Settings}: We set the hidden size of the convolutional layer to $200$, the dropout rate among $\{0.43, 0.5\}$, the learning rate among $\{0.0018, 0.018\}$, the weight decay among $\{0,8e-6\}$, the maximum training epoch with Adam to 1000 without early stopping. The parameter settings in all baseline models are the same as \cite{kipf2017semi,simplifyGCN}.

\textbf{Performance}: As shown in Table \ref{tab:citation-results}, our proposed method achieves the best performance over all baseline methods on Citeseer and Pubmed datasets. It shows that our model can be well generalized to test data using only document citation information in training data. Compared with baseline methods, besides the given citation network, our model  incorporates some additional inter-document correlation information derived from document-word features and word-word graph, which could benefit our final  performance. 

\begin{table}[!hbt]\footnotesize
    \centering
    \renewcommand{\arraystretch}{1.2}
    \caption{Test Accuracy (\%) averaged over 10 runs on citation networks. WGCN steadily outperforms other methods on Citeseer and Pubmed based on student t-test ($p-value <0.05$). }
    \begin{tabular}{l|c|c}
    \hline
    \bf{Model}      & \bf{Citeseer}             & \bf{Pubmed}    \\ 
    \hline
    GCN           & 70.3 & 79.0 \\
    Fast-GCN      & 68.8 $\pm$ 0.6 & 77.4 $\pm$ 0.3  \\
    % SAGE-GCN     & - & - 8 - & 90.8 \\
    GIN     & 70.9 $\pm$ 0.1 & 78.9 $\pm$ 0.1 \\
    SGC          & 71.9 $\pm$ 0.1 & 78.9 $\pm$ 0.0 \\ % & 94.9 \\
    %WGCN-mask         &  72.2 $\pm$ 0.1  & 79.6 $\pm$ 0.0   \\
	WGCN         &  \textbf{72.2} $\pm$ 0.1  & \textbf{79.6} $\pm$ 0.0   \\
    \hline
    \end{tabular}
    \label{tab:citation-results}
\end{table}

In Table \ref{tab:citation-A-results}, we further present the test accuracy of our method with different orders of neighborhood information of $A_{dd}$. We can see that our method achieves the best performance when considering 1-order neighborhood information on both datasets. Considering 0-order citation information, i.e. the inter-document relationships are only calculated from document-word features, may not be sufficient for information propagation, while more than 1-order information may over-propagate to moderately related documents, 
%which leads to over-smoothing representation, 
resulting in worse performance in both scenarios. 
\begin{table}[!hbt]\footnotesize
    \centering
    \renewcommand{\arraystretch}{1.2}
    \caption{Test Accuracy (\%) averaged over 10 runs on citation networks using different order neighborhood information of $A_{dd}$, i.e. $A_{dd}^k, k\in(0,1,2,3)$. }
    \begin{tabular}{c|c|c}
    \hline
    \bf{Model}      & \bf{Citeseer}             & \bf{Pubmed}    \\ 
    \hline
	$A_{dd}^0$ &  70.5 $\pm$ 0.0  & 78.0 $\pm$ 0.0   \\
	$A_{dd}^1$ &  \textbf{72.2} $\pm$ 0.1  & \textbf{79.6} $\pm$ 0.0   \\
	$A_{dd}^2$ &  67.2 $\pm$ 0.1  & 78.9 $\pm$ 0.0   \\
	$A_{dd}^3$ &  64.8 $\pm$ 0.1  & 78.4 $\pm$ 0.0   \\
    \hline
    \end{tabular}
    \label{tab:citation-A-results}
\end{table}

%%%%%%%%%%%%%%%%%%%%%%%%%%%%%%%%%
%%         Conclusion          %%
%%%%%%%%%%%%%%%%%%%%%%%%%%%%%%%%%

\section{Conclusion}
%In this paper, we introduced a novel inductive GCN model to learn document representations in supervised manner, which can be generalized to out-of-graph documents. The key component is to degrade the document-level graph into a word-level graph, to decouple data samples and GCN models with a document-independent graph. Our analysis provides insight into how our method connecting with existing methods. Experimental results demonstrate the efficiency and effectiveness of our method along with the advantage of decoupling the data samples from the adjacency graph. For  future work, it would be interesting to explore how to learn embedding for documents which contains unseen words in WGraph, to cope with out of vocabulary (OOV) problem. It is also possible to extend our model to unsupervised representation learning with a large amount of unlabeled data.

In this paper, we introduced a novel inductive GCN model to learn document representations in a supervised manner. The advantages of the proposed  WGCN model are twofold. Firstly,  WGCN degrades the document-level graph into a word-level graph,  thereby reducing the complexity of the graph especially when the document number is large than the size of word vocabulary in real-world applications. As a result, the efficiency of convolution operations become much more efficient.
Plus, WGCN transforms transductive learning to inductive learning, since it  decouples data samples and GCN models with a document-independent graph. Therefore, WGCN  can be generalized to out-of-graph documents, i.e., the documents to be classified do not need to be included in the graph when the model is trained.  This is crucial for real-world applications.

%Our analysis provides insight into how our method connecting with existing methods.
Experimental results demonstrated the efficiency and effectiveness of our method along with the advantage of decoupling the data samples from the adjacency graph. For  future work, it would be interesting to explore how to learn representations for documents which contain unseen words in WGraph, to cope with out of vocabulary (OOV) problem. It is also possible to extend our model to unsupervised representation learning with a large amount of unlabeled data.

\newpage

\bibliography{acmart.bib}

\end{document}